\documentclass[11pt]{article}
\usepackage[margin=1in]{geometry}

\usepackage{booktabs,threeparttable}
\usepackage{graphicx}
\usepackage{amsmath}
\usepackage{hyperref}
\usepackage{float} 
\usepackage{bookmark} 
\usepackage{adjustbox}

\AtBeginDocument{%
  }

\begin{document}

\title{Hierarchical Section Matching Prediction (HSMP) BERT for Fine-Grained Extraction of Structured Data from Hebrew Free-Text Radiology Reports in Crohn’s Disease}

\author{
Zvi Badash\textsuperscript{1},
Hadas Ben-Atya\textsuperscript{2},
Naama Gavrielov\textsuperscript{2},
Liam Hazan\textsuperscript{1},\\
Gili Focht\textsuperscript{3},
Ruth Cytter-Kuint\textsuperscript{4},
Talar Hagopian\textsuperscript{4},
Dan Turner\textsuperscript{3},\\
Moti Freiman\textsuperscript{2}\thanks{Corresponding author. Email: moti.freiman@technion.ac.il}
}

\date{
\textsuperscript{1}Faculty of Data and Decision Sciences, Technion -- Israel Institute of Technology, Haifa, Israel\\
\textsuperscript{2}Faculty of Biomedical Engineering, Technion -- Israel Institute of Technology, Haifa, Israel\\
\textsuperscript{3}Juliet Keidan Institute of Pediatric Gastroenterology, Hepatology and Nutrition,\\
The Eisenberg R\&D Authority, Shaare Zedek Medical Center, The Hebrew University School of Medicine, Jerusalem, Israel\\
\textsuperscript{4}Pediatric Radiology Unit, Radiology Department,\\
The Eisenberg R\&D Authority, Shaare Zedek Medical Center, The Hebrew University of Jerusalem, Jerusalem, Israel
}

\maketitle

\begin{abstract}
  Extracting structured clinical information from radiology reports is challenging, especially in low-resource languages.  This is pronounced in Crohn’s disease, with sparsely represented multi-organ findings. We developed Hierarchical Structured Matching Prediction BERT (HSMP-BERT), a prompt-based model for extraction from Hebrew radiology text. In an administrative database study, we analyzed 9,683 reports from Crohn’s patients imaged 2010–2023 across Israeli providers. A subset of 512 reports was radiologist-annotated for findings across six gastrointestinal organs and 15 pathologies, yielding 90 structured labels per subject. Multilabel-stratified split (66\% train+validation; 33\% test), preserving label prevalence. Performance was evaluated with accuracy, F1, Cohen’s $\kappa$, AUC, PPV, NPV, and recall. On 24 organ–finding combinations with $>$15 positives, HSMP-BERT achieved mean F1 0.83$\pm$0.08 and $\kappa$ 0.65$\pm$0.17, outperforming the SMP zero-shot baseline (F1 0.49$\pm$0.07, $\kappa$ 0.06$\pm$0.07) and standard fine-tuning (F1 0.30$\pm$0.27, $\kappa$ 0.27$\pm$0.34; paired t-test  $p < 10^{-7}$). Hierarchical inference cuts runtime 5.1$\times$ vs. traditional inference. Applied to all reports, it revealed associations among ileal wall thickening, stenosis, and pre-stenotic dilatation, plus age- and sex-specific trends in inflammatory findings. HSMP-BERT offers a scalable solution for structured extraction in radiology, enabling population-level analysis of Crohn’s disease and demonstrating AI’s potential in low-resource settings.

\end{abstract}

\section{Introduction}
Radiology reports contain rich clinical signal central to patient care, research, and decision support, yet their unstructured, narrative form hinders systematic, large-scale analysis~\cite{Reschke2025}. Manual abstraction is time-consuming, error-prone, and infeasible in high-throughput environments~\cite{Jorg2023}. Advances in natural language processing (NLP)~\cite{Mozayan2021,Nowak2023,Graf2025,Wonicki2025,Schlegel2020}, particularly large language models (LLMs), enable conversion of free text into structured representations for disease surveillance, cohort curation, and AI model development~\cite{Reichenpfader2024,Guellec2024,Liu2025,TayebiArasteh2025,Linguraru2024}.

Despite progress, deploying LLMs in clinical workflows remains challenging~\cite{Artsi2025}. Practical constraints include data privacy (e.g., reliance on external inference endpoints), limited support for non‑English and low-resource languages~\cite{Dennstdt2025,Busch2025,Ong2024,AkinciDAntonoli2025}, and the domain shift from general-purpose pretraining to clinical terminology—especially acute in Hebrew, where annotated corpora are scarce~\cite{Malul2025,Sha2025}. Clinical text also exhibits marked class imbalance: rare yet critical findings are underrepresented, complicating robust learning and calibration~\cite{Li2025,Avnat2025}.

Crohn’s disease exemplifies these issues. Presentations are heterogeneous across gastrointestinal organs, and radiologic descriptions may be subtle or infrequent (e.g., wall thickening, enhancement, stenosis, pre‑stenotic dilatation)~\cite{bruining2018consensus}. Accurately structuring such findings is necessary for population health monitoring and research but technically demanding at scale.

Prompt-based learning provides a pragmatic path for low-resource clinical NLP by eliciting task behavior with minimal labeled data~\cite{Jiang2022,Zhuang2025,Reichenpfader2025}. Moreover, document structure can be leveraged as self-supervision: radiology reports typically encode a logical relationship between \emph{Findings} and \emph{Impression}~\cite{Hazan2024}.

In this work, we present \emph{Hierarchical Section Matching Prediction BERT (HSMP‑BERT)}, a prompt-based framework for fine-grained extraction of organ–finding labels from Hebrew radiology reports. HSMP‑BERT combines (i) a \emph{Section Matching Prediction} (SMP) objective that exploits Findings–Impression consistency for domain-aware pretraining with (ii) a \emph{hierarchical inference} scheme that routes from study-level to organ-level to finding-level decisions. We evaluate the approach on a national cohort of inflammatory bowel disease (IBD) patients from Israel, comprising 9,683 MRE/CTE reports (2010–2023) and an expert-annotated subset of 512 reports covering six organs and 15 pathologies, yielding 90 labels per subject~\cite{Friedman2018}.

\paragraph*{Contributions}
\begin{itemize}
  \item We introduce HSMP‑BERT, which unites section‑aware pretraining (SMP) with hierarchical prompting/inference for structured extraction in a low‑resource clinical language.
  \item We provide a reproducible evaluation for 24 organ–finding targets with sufficient prevalence, reporting predictive performance, calibration, and efficiency; hierarchical routing substantially reduces inference cost while preserving accuracy.
  \item We apply the validated model at scale to characterize clinically meaningful patterns by age, sex, and anatomical site, illustrating utility for population-level analyses in Crohn’s disease.
\end{itemize}

\paragraph*{Paper roadmap} Section~\ref{sec:methods} describes the dataset, annotation protocol, SMP pretraining, and hierarchical inference. Section~\ref{sec:results} reports predictive performance, calibration, and efficiency. Section~\ref{sec:discussion} discusses clinical relevance, limitations (e.g., domain shift, class imbalance, label recoding), and implications for deployment in low-resource settings. Finally, Section~\ref{sec:conclusion} concludes the paper.

\section{Materials and methods}
\label{sec:methods}
\subsection{Data Collection}
\label{sec:data}

This multicenter administrative study was approved by the relevant institutional review board, with a waiver of informed consent due to its retrospective nature, the de-identification of data, and the large cohort size.

Hebrew-language free-text radiology reports were obtained from the \textit{epi-IIRN} national inflammatory bowel disease (IBD) study cohort, a validated database encompassing all four Israeli health maintenance organizations (HMOs), which collectively cover over 98\% of the population~\cite{Friedman2018}. The dataset comprises \textbf{9,683 radiology reports} corresponding to individual patient visits, derived from \textbf{8,093 unique patients}. Each report describes either a \textit{magnetic resonance enterography} (MRE) or \textit{computed tomography enterography} (CTE) examination, performed between \textbf{2010 and 2023} at one of several participating imaging providers across Israel.

\subsubsection{Manual Annotation of Organ--Finding Pairs}

A random subset of \textbf{512 reports} was selected for expert manual annotation. Each report was labeled independently by two board-certified radiologists (initial labeling by R1, reviewed and adjudicated by R2; names withheld for anonymization). Annotations were structured across:

\begin{itemize}
  \item \textbf{Six gastrointestinal organs}: Jejunum, Ileum, Cecum, Colon, Sigmoid, Rectum
  \item \textbf{Up to 15 pathologies per organ}, including: wall thickening, lumen narrowing, wall enhancement, ulceration, fistula, abscess, mesenteric edema, reduced motility, DWI signal abnormalities, and others
\end{itemize}

This resulted in a total of \textbf{90 possible organ--finding combinations} per report. Each combination was assigned one of four categorical labels:

\begin{itemize}
  \item 1 = finding present (positive)
  \item 0 = finding absent (negative)
  \item 2 = organ not visible (e.g., outside scan field)
  \item 9 = organ resected (post-surgical)
\end{itemize}

For the main classification task, labels 2 and 9 were treated as negative, simplifying the task to a \textbf{binary classification} of finding presence vs. absence. While this design choice improves tractability, its effects are analyzed further in Section~\ref{sec:limitations}.

\begin{figure}[t!]
    \centering
    \includegraphics[width=0.9\textwidth]{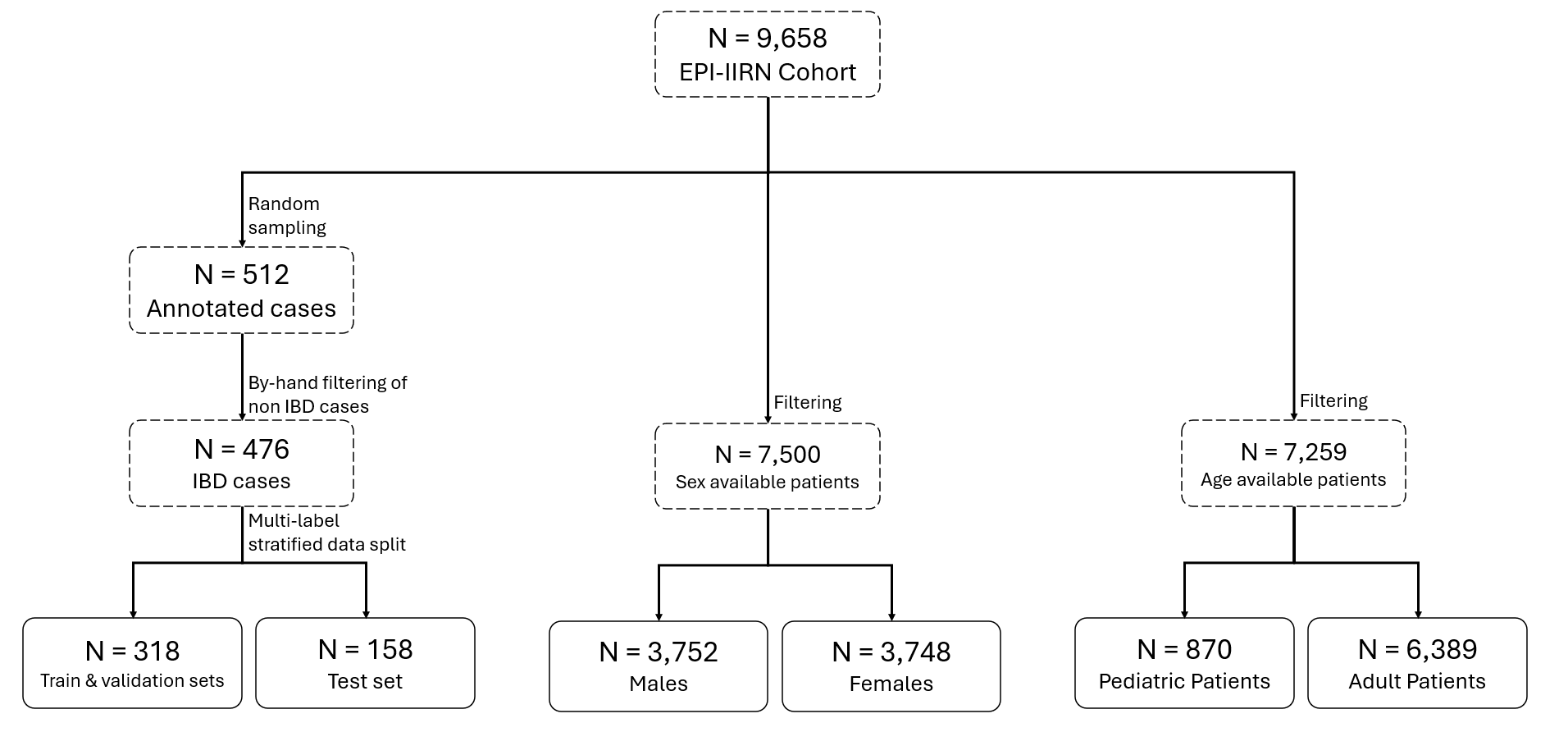}
    \caption{Data selection flowchart. The leftmost arm of the diagram shows how radiology reports were chosen for annotation and the multi-label stratified split. The middle and rightmost arms describe the population's large-scale analysis of disease was performed on.}
    \label{fig:data_selection}
\end{figure}

To ensure sufficient positive cases for evaluation, we retained only those organ--finding pairs with at least 15 positive examples, resulting in \textbf{24 final classification targets}. The annotation and filtering process is summarized in Figure~\ref{fig:data_selection}, and the full schema of organs and findings appears in Table~\ref{tab:organs_findings}.

\begin{table}[t!]
\centering
\caption{Organs and findings annotated in our dataset and later used for disease analysis on a large scale.}
\begin{tabular}{|l|l|l|}
\hline
\textbf{Organs} & \multicolumn{2}{c|}{\textbf{Findings}} \\ \hline
Jejunum        & Inflammation        & Phlegmon \\ \hline
Ileum          & Fistula             & Mesenteric edema (or fat stranding) \\ \hline
Cecum          & Stenosis            & DWI signal \\ \hline
Colon          & Pseudosacculation   & Abscess \\ \hline
Sigmoid colon  & Comb sign           & Wall thickness \\ \hline
Rectum         & Sinus tract         & Ulcer \\ \hline
               & Bowel wall edema    & Wall enhancement \\ \hline
               & Pre-stenotic dilation & Reduced motility \\ \hline
\end{tabular}
\label{tab:organs_findings}
\end{table}

\subsection{HSMP-BERT: Pretraining, Prompting, and Inference}
To support structured information extraction from Hebrew radiology reports, we developed HSMP-BERT by extending the Hebrew RoBERTa (HeRo) model~\cite{Shalumov2023}, originally pre-trained on the HeDC4 corpus. Given the scarcity of publicly available Hebrew medical corpora, we adapted HeRo to the radiology domain through additional pretraining on all 9,683 MRE and CTE reports described in Section~\ref{sec:data}. This domain-adaptive step used the Masked Language Modeling (MLM) objective~\cite{Gururangan2020} to familiarize the model with clinical vocabulary, abbreviations, and report structure.

\subsubsection{Section Matching Prediction Pretraining}

To further specialize the model for structured inference, we introduced a novel auxiliary task: \textit{Section Matching Prediction} (SMP), which exploits the natural structure of radiology reports~\cite{Hazan2024}. Most reports contain a \textit{Findings} section (detailed image observations) and an \textit{Impression} section (interpretive summary). We frame SMP as a binary classification task to determine whether a pair of Findings and Impression sections originates from the same report (\texttt{Match}) or from unrelated ones (\texttt{NotMatch}), analogous to Next Sentence Prediction.

During SMP training, we construct input pairs as:
\begin{equation*}
  x_{\text{in}} = \texttt{[CLS]} ~ x^F_i ~ \texttt{[SEP]} ~ x^I_i ~ \texttt{[EOS]}
\end{equation*}
where $x^F_i$ and $x^I_i$ represent the tokenized \textit{Findings} and \textit{Impression} sections, respectively. The model computes a hidden representation $h_{\texttt{[CLS]}}$, which is passed to a classifier:
\begin{equation*}
  q_\mathcal{M}(n_k | x_i) = \texttt{softmax}(W_{\texttt{smp}} h_{\texttt{[CLS]}} + b_{\texttt{smp}})
\end{equation*}
with $n_k \in \{\texttt{Match}, \texttt{NotMatch}\}$.

This pretraining objective enables HSMP-BERT to internalize the logical structure of radiology reports, a critical prerequisite for label-efficient extraction tasks in low-resource languages like Hebrew~\cite{Hazan2024}.

\subsubsection{Prompt Construction and SMP-Tuning}

Following SMP pretraining, we designed a prompting mechanism for structured extraction using a \textit{verbalizer} function \( f: \mathcal{Y} \to \mathcal{P} \), which maps each label $y_j$ (e.g., ``wall thickening in ileum'') to a templated natural language prompt $p_j$. Prompts were designed to mirror clinical phrasing and support inference without additional layers. Each input took the form:
\begin{equation*}
  x_{\text{in}} = \texttt{[CLS]} ~ x^F_i ~ \texttt{[SEP]} ~ p_j ~ \texttt{[EOS]}
\end{equation*}
Example prompt template:
\begin{equation*}
  \mathcal{T}(x_i) = \texttt{[CLS]} ~ x^F_i ~ \texttt{[SEP]} ~ \text{``There is \{finding\} in the \{organ\}.''} ~ \texttt{[EOS]}
\end{equation*}
This structure allows HSMP-BERT to predict the compatibility of each organ--finding pair with the report content.

We refer to the process of refining the SMP-trained model on annotated examples using these templates as \textit{SMP-tuning}. For each gold label $y^+_i$, we construct a positive instance $(\mathcal{T}(x_i, y^+_i), 1)$. For each incorrect label $y^- \in \mathcal{Y} \setminus \{y^+_i\}$, we construct negative instances $(\mathcal{T}(x_i, y^-), 0)$. We fine-tune the model using a binary cross-entropy loss.

\subsubsection{Hierarchical Prompted Inference}
\label{sec:prompted_inference}
To improve inference efficiency, we applied a \textit{hierarchical prompting} strategy. Instead of evaluating all organ--finding combinations exhaustively, we structured inference as a decision tree with three levels:
\begin{enumerate}
  \item Scan-level prompt: ``Is the scan normal?''
  \item Organ-level prompt: e.g., ``Is there any abnormality in the ileum?''
  \item Finding-level prompt: e.g., ``Is there bowel wall thickening in the ileum?''
\end{enumerate}

Queries proceed top-down: if a higher-level prompt yields a negative prediction, lower-level prompts for that subtree are skipped. This reduces the number of calls to the model and total tokens processed, yielding significant gains in runtime without sacrificing accuracy.

\begin{figure}[t!]
    \centering
    \includegraphics[width=0.9\textwidth]{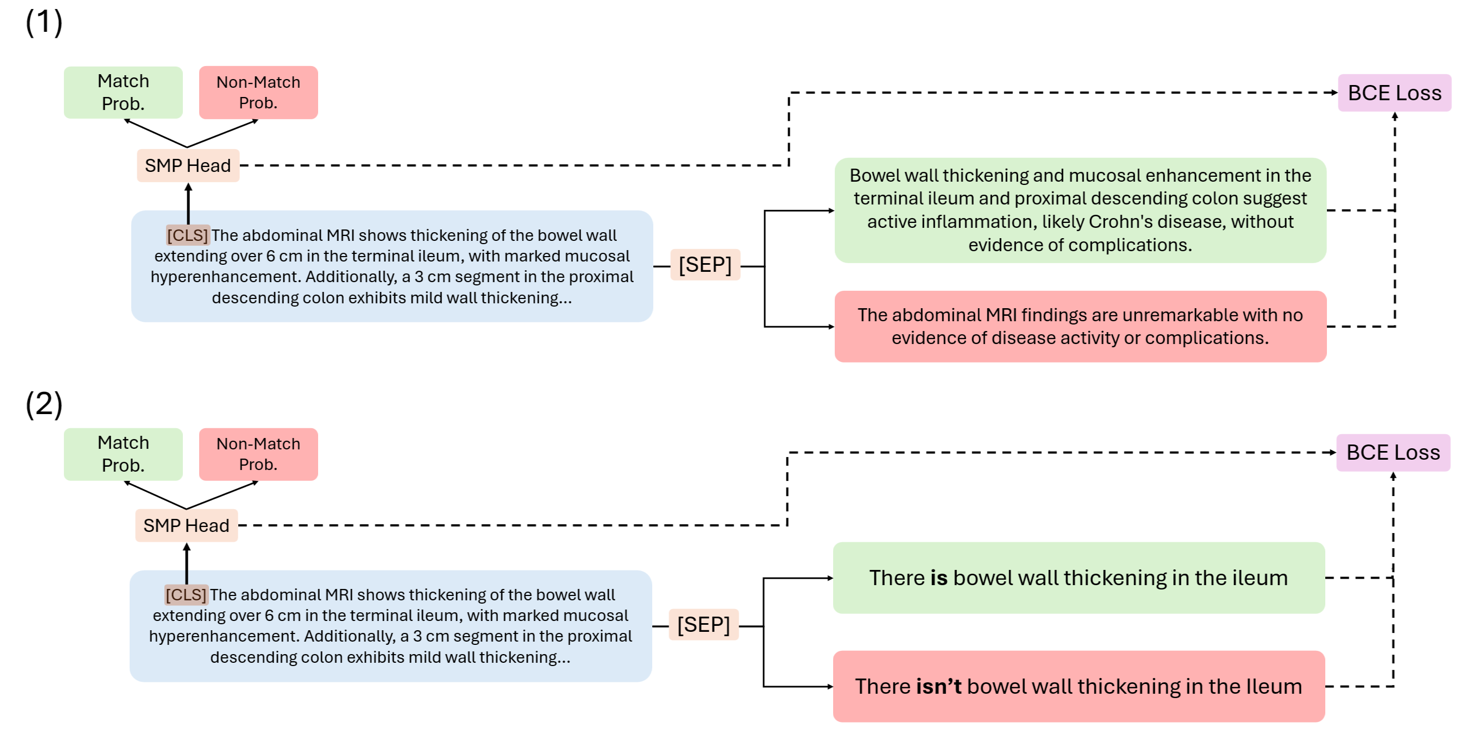}
    \caption{Overview of the Hierarchical SMP-BERT training framework for structured data extraction from free-text radiology reports. (1) Section Matching Prediction (SMP) Pre-training: The model learns to distinguish matching vs. non-matching pairs of “Findings” and “Impression” sections from radiology reports using a binary classification head and BCE loss. (2) Fine-tuning: Using annotated reports, the model is optimized to predict the presence or absence of specific clinical findings (e.g., bowel wall thickening in the ileum) based on matched or mismatched impressions.}
    \label{fig:hsmp_training}
\end{figure}

\begin{figure}[t!]
    \centering
    \includegraphics[width=0.9\textwidth]{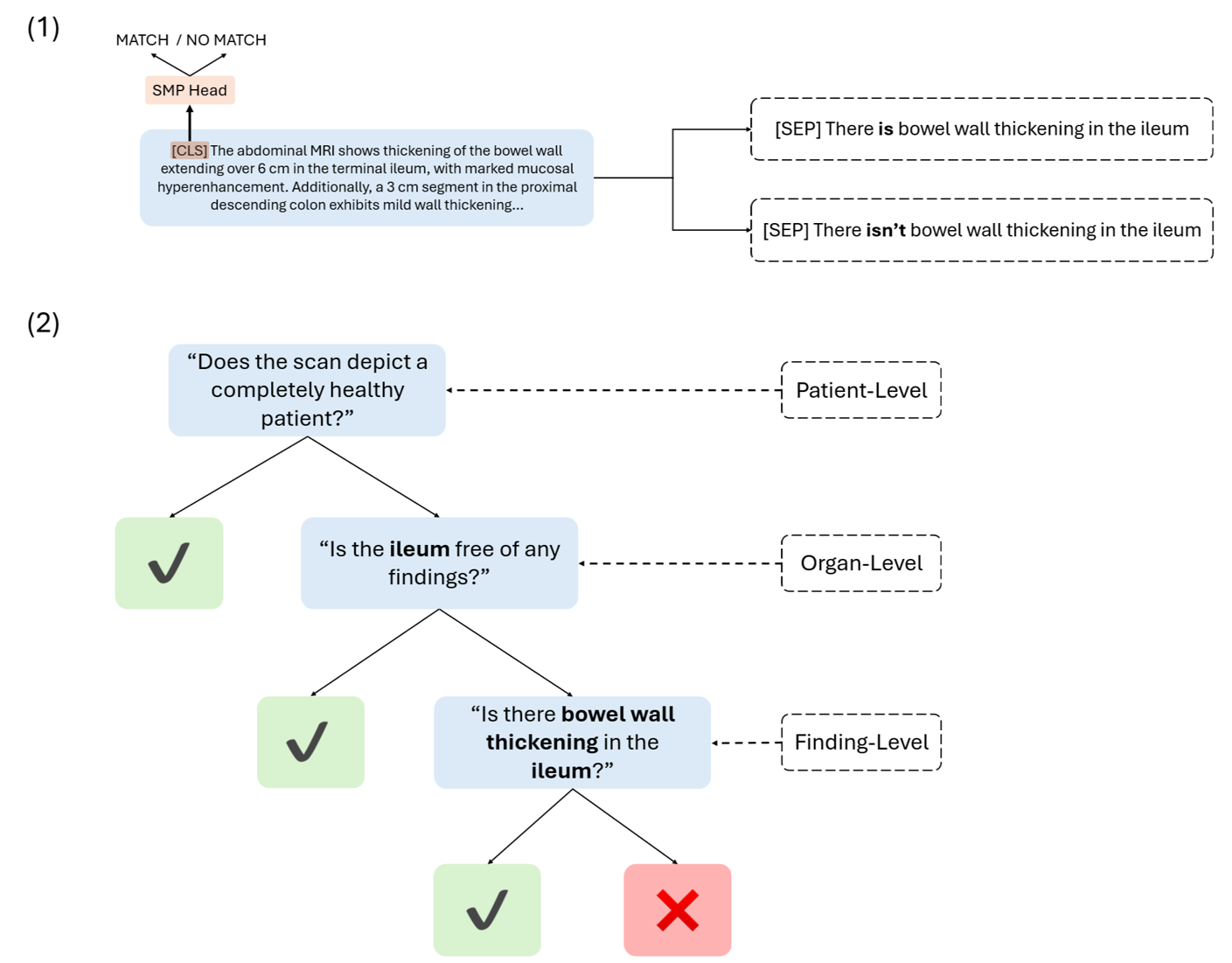}
    \caption{Overview of the Hierarchical SMP-BERT inference framework for structured data extraction from free-text radiology reports. (1) Traditional inference: At test time, the model is prompted with structured clinical hypotheses and determines whether the report supports each statement. (2) Hierarchical Inference: A tree-based querying strategy reduces inference steps by prioritizing general questions (e.g., "Is the scan normal?") before proceeding to more specific findings, improving scalability and efficiency.}
    \label{fig:hsmp_inference}
\end{figure}

Figures~\ref{fig:hsmp_training} and~\ref{fig:hsmp_inference} illustrate the training and inference architecture of HSMP-BERT.

\subsection{Evaluation Metrics and Statistical Analysis}
\subsection{Evaluation Framework}

We evaluated the performance of HSMP-BERT using standard metrics for multi-label clinical classification, including F1 score, area under the receiver operating characteristic curve (AUC), precision, recall, Cohen’s $\kappa$, positive predictive value (PPV), and negative predictive value (NPV). These metrics were computed per organ--finding label and averaged across the 24 tasks retained for analysis (see Section~\ref{sec:data}).

To assess the contribution of each modeling component, we compared three configurations:

\begin{itemize}
  \item \textbf{Zero-shot HSMP-BERT:} uses the pretrained model without supervised fine-tuning.
  \item \textbf{Fine-tuned HSMP-BERT:} builds on the SMP-pretrained model with further tuning on the manually labeled subset.
  \item \textbf{Standard fine-tuning (SFT):} directly trains a classification model on the labeled subset without SMP pretraining or prompting.
\end{itemize}

We additionally compared two inference strategies:

\begin{itemize}
  \item \textbf{Flat inference:} exhaustively evaluates all organ--finding prompts.
  \item \textbf{Hierarchical inference:} employs a top-down query routing approach as described in Section~\ref{sec:prompted_inference}.
\end{itemize}

To ensure consistent label distributions across sets, we performed a multilabel-stratified split of the annotated data (66\% for training and validation; 34\% for held-out evaluation). Each model was trained and evaluated using identical splits to support direct comparison. All evaluations were conducted using five random seeds, and we report mean and standard deviation unless otherwise specified.

Beyond supervised classification, we evaluated HSMP-BERT in a deployment-style setting by applying the trained model to the full corpus of 9,683 radiology reports. This allowed us to conduct population-level analyses, including:

\begin{itemize}
  \item \textbf{Stratified prevalence estimation:} comparing finding frequencies by sex and age group (adults: $>$18 years; pediatric: $\leq$18 years).
  \item \textbf{Organ--finding correlation analysis:} computing pairwise associations to identify co-occurrence patterns (e.g., wall thickening and pre-stenotic dilatation).
\end{itemize}

To assess generalizability, we validated consistency between predictions on the full dataset and the annotated subset. Specifically, we compared aggregate patterns (e.g., prevalence, rank-order, correlation structure) between model-inferred and expert-annotated data. Agreement between these distributions supports the model’s reliability for downstream applications in clinical informatics, population surveillance, and retrospective research.

\section{Results}
\label{sec:results}
\begin{figure}[t!]
    \centering
    \includegraphics[width=0.9\textwidth]{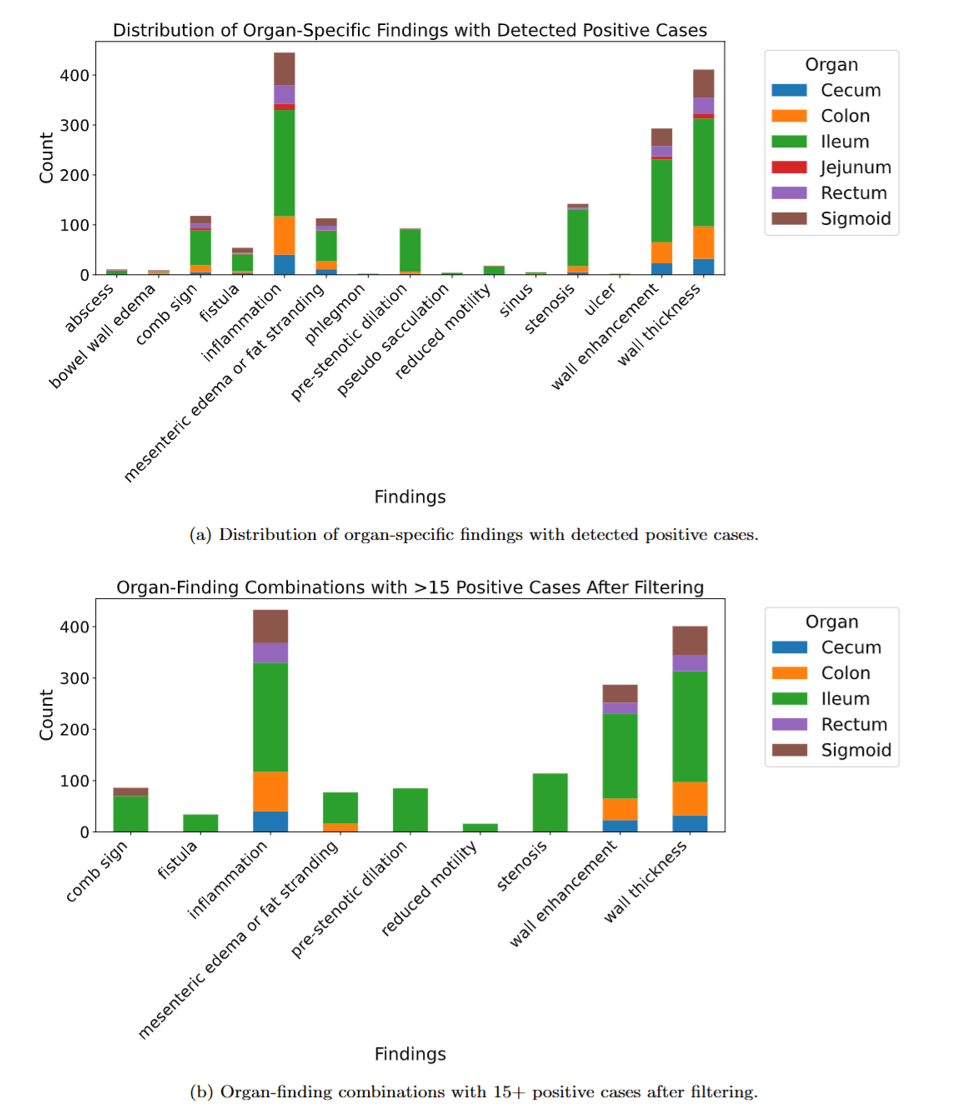}
    \caption{(a) Distribution of organ-specific findings and (b) filtered organ-finding combinations with more than 15 positive cases.}
    \label{fig:distribution}
\end{figure}

The dataset used in this study comprised 9,658 free-text radiology reports from 7,389 unique patients diagnosed with Crohn’s disease, collected between 2010 and 2023. These reports were sourced from four of Israel’s largest health maintenance organizations (HMOs), covering approximately 98\% of the national population. The mean patient age was 37 years (range: 11-78 yrs), and the cohort included 50\% male and 50\% female patients. Figure~\ref {fig:distribution} shows the distribution of all organ-specific findings and the subset with at least 15 positive samples. Detailed information about the prevalence of the different labels in the test set is provided in Table~\ref{tab:supp_label_prevalence} in the supplementary materials. 

Table~\ref{tab:hsmp-bert-performance} summarizes the performance of the HSMP-BERT model across 24 organ–finding combinations extracted from manually annotated radiology reports. Overall, the model demonstrated high accuracy and generalizability across both common and rare clinical findings, achieving excellent results on key evaluation metrics including accuracy, F1 score, Cohen’s $\kappa$, balanced accuracy, and positive predictive value (PPV) and negative predictive value (NPV). Tables~\ref{tab:s2a_metrics},~\ref{tab:s2b_metrics}, and~\ref{tab:s2c_metrics} present the performance metrics for the baseline models, i.e. fine-tuned SMP-BERT, zero-shot SMP-BERT, SFT-BERT.  

\begin{table}[t!]
\centering
\caption{Performance metrics of the HSMP-BERT model across 24 organ–finding combinations in annotated Crohn’s disease radiology reports.}
\resizebox{\textwidth}{!}{%
\begin{tabular}{|l|c|c|c|c|c|c|}
\hline
\textbf{Label} & \textbf{Accuracy} & \textbf{F1 Score} & \textbf{Cohen's $\kappa$} & \textbf{Balanced Accuracy} & \textbf{NPV} & \textbf{PPV} \\ \hline
Cecum Inflammation       & 0.96 & 0.84 & 0.67 & 0.85 & 0.98 & 0.67 \\ \hline
Cecum Wall Enhancement   & 0.97 & 0.83 & 0.65 & 0.78 & 0.98 & 0.80 \\ \hline
Cecum Wall Thickness     & 0.96 & 0.84 & 0.68 & 0.84 & 0.98 & 0.70 \\ \hline
Colon Inflammation       & 0.94 & 0.88 & 0.77 & 0.90 & 0.97 & 0.78 \\ \hline
Colon Wall Enhancement   & 0.94 & 0.79 & 0.58 & 0.77 & 0.96 & 0.67 \\ \hline
Colon Wall Thickness     & 0.93 & 0.85 & 0.69 & 0.83 & 0.95 & 0.79 \\ \hline
Ileum Comb sign          & 0.91 & 0.80 & 0.60 & 0.77 & 0.93 & 0.76 \\ \hline
Ileum DWI Signal         & 0.97 & 0.93 & 0.87 & 0.91 & 0.97 & 0.95 \\ \hline
Ileum Fistula            & 0.98 & 0.93 & 0.87 & 0.99 & 1.00 & 0.79 \\ \hline
Ileum Inflammation       & 0.87 & 0.87 & 0.73 & 0.88 & 0.94 & 0.79 \\ \hline
Ileum Mesenteric Edema   & 0.84 & 0.65 & 0.30 & 0.65 & 0.91 & 0.38 \\ \hline
Ileum Pre Stenotic Dilatation & 0.94 & 0.91 & 0.81 & 0.91 & 0.97 & 0.83 \\ \hline
Ileum Reduced Motility   & 0.98 & 0.78 & 0.56 & 0.70 & 0.98 & 1.00 \\ \hline
Ileum Stenosis           & 0.95 & 0.93 & 0.87 & 0.94 & 0.97 & 0.87 \\ \hline
Ileum Wall Enhancement   & 0.93 & 0.93 & 0.86 & 0.94 & 0.98 & 0.87 \\ \hline
Ileum Wall Thickness     & 0.92 & 0.92 & 0.85 & 0.93 & 0.97 & 0.87 \\ \hline
Rectum Inflammation      & 0.94 & 0.80 & 0.61 & 0.85 & 0.98 & 0.56 \\ \hline
Rectum Wall Enhancement  & 0.96 & 0.80 & 0.61 & 0.84 & 0.99 & 0.66 \\ \hline
Rectum Wall Thickness    & 0.97 & 0.88 & 0.77 & 0.94 & 0.99 & 0.69 \\ \hline
Sigmoid Comb sign        & 0.97 & 0.63 & 0.27 & 0.60 & 0.97 & 0.52 \\ \hline
Sigmoid Inflammation     & 0.87 & 0.75 & 0.51 & 0.79 & 0.95 & 0.52 \\ \hline
Sigmoid Mesenteric Edema & 0.97 & 0.74 & 0.49 & 0.70 & 0.98 & 0.67 \\ \hline
Sigmoid Wall Enhancement & 0.94 & 0.77 & 0.54 & 0.76 & 0.97 & 0.60 \\ \hline
Sigmoid Wall Thickness   & 0.89 & 0.74 & 0.48 & 0.75 & 0.94 & 0.53 \\ \hline
\end{tabular}
}
\label{tab:hsmp-bert-performance}
\end{table}

Among the most prevalent findings, the model performed especially well on ileum wall thickness (45\% prevalence), with an F1 score of 0.93, Cohen’s $\kappa$ of 0.86, and balanced accuracy of 0.94. Similarly, Ileum inflammation (41\% prevalence) achieved an F1 of 0.87 and Cohen’s $\kappa$ of 0.73. Despite the inherent variability in report language and imaging descriptions, the model consistently reached high agreement with expert annotations.
The model also maintained strong performance for less frequent findings. For example, rectum wall thickness (6\% prevalence) achieved an F1 score of 0.88, Cohen’s $\kappa$ of 0.77, and a near-perfect NPV of 0.9. Rare findings such as sigmoid comb sign and rectal wall enhancement (with prevalences 3\% and 7\%, respectively) achieved F1 scores of 0.63 and 0.80, respectively, underscoring the model’s ability to generalize well even in low-resource, low-prevalence scenarios.
A repeated-measures ANOVA revealed a significant effect of model on F1 score and Cohen’s $\kappa$ across labels (p $<$ 10$^{-20}$). Post-hoc pairwise t-tests (Benjamini-Hochberg-corrected) showed that HSMP-BERT significantly outperformed SMP Zero-Shot ($\Delta$F1 = 0.33, p $<$ 10$^{-15}$ and $\Delta \kappa$ = 0.59, p \textless 10$^{-14}$) and SFT ($\Delta$F1 = 0.53, p \textless 10$^{-8}$ and $\Delta \kappa$ = 0.38, p \textless 10$^{-7}$). No significant difference was observed between HSMP-BERT and SMP Finetune (p $>$ 0.80).

The hierarchical inference, which structures queries from general to specific, led to a 5.1$\times$ speedup in inference time compared to vanilla inference (1.77 vs. 0.35 seconds per single radiology report, on average). Table~\ref{tab:inference_efficiency} describes the speedup in more detail.

\begin{table}[t!]
\centering
\caption{Comparison of inference efficiency between flat (SMP-BERT) and hierarchical (HSMP-BERT) inference schemes in terms of total runtime, number of BERT forward calls, and total tokens processed. The presented numbers are summed over all 9,658 samples in the dataset. All metrics show substantial reductions under the hierarchical setup, indicating improved computational efficiency.}
\begin{tabular}{|l|c|c|c|}
\hline
\textbf{Model} & \textbf{Inference Time (hh:mm:ss)} & \textbf{BERT Calls} & \textbf{Tokens} \\ \hline
SMP-BERT       & 4:46:00 & 2,180,436 & 574,351,210 \\ \hline
HSMP-BERT      & 0:55:36 &   395,750 & 113,743,598 \\ \hline
Improvement (fold) & 5.14x & 5.51x & 5.05x \\ \hline
\end{tabular}
\label{tab:inference_efficiency}
\end{table}

Table~\ref{tab:avg_metrics} compares the average metrics over the different labels for the different models. HSMP-BERT and SMP-Finetune, achieved the same accuracy, but HSMP-BERT reduced the computational cost substantially. Both models performed better in all metrics compared to SMP zero-shot and standard SFT training.

\begin{table}[t!]
\centering
\caption{The average metrics over the different labels for the different models. HSMP-BERT and SMP-Finetune achieved the same accuracy, but HSMP-BERT reduced the computational cost substantially. Both models performed better in all metrics compared to SMP zero-shot and standard SFT training.}
 \resizebox{\textwidth}{!}{%
\begin{tabular}{|l|c|c|c|c|c|c|}
\hline
\textbf{Method} & \textbf{Accuracy} & \textbf{F1 Score} & \textbf{Cohen’s $\kappa$} & \textbf{Balanced Accuracy} & \textbf{NPV} & \textbf{PPV} \\ \hline
HSMP-BERT     & 0.94 $\pm$ 0.04 & 0.83 $\pm$ 0.08 & 0.65 $\pm$ 0.17 & 0.83 $\pm$ 0.10 & 0.97 $\pm$ 0.02 & 0.71 $\pm$ 0.16 \\ \hline
SMP Finetune  & 0.94 $\pm$ 0.04 & 0.83 $\pm$ 0.08 & 0.65 $\pm$ 0.17 & 0.83 $\pm$ 0.10 & 0.97 $\pm$ 0.02 & 0.71 $\pm$ 0.15 \\ \hline
SMP Zero-shot & 0.65 $\pm$ 0.08 & 0.49 $\pm$ 0.07 & 0.06 $\pm$ 0.09 & 0.55 $\pm$ 0.07 & 0.88 $\pm$ 0.10 & 0.18 $\pm$ 0.15 \\ \hline
SFT           & 0.92 $\pm$ 0.03 & 0.30 $\pm$ 0.34 & 0.27 $\pm$ 0.31 & 0.62 $\pm$ 0.15 & 0.93 $\pm$ 0.03 & 0.65 $\pm$ 0.31 \\ \hline
\end{tabular}
}
\label{tab:avg_metrics}
\end{table}

\begin{figure}[t!]
    \centering
    \includegraphics[width=0.9\textwidth]{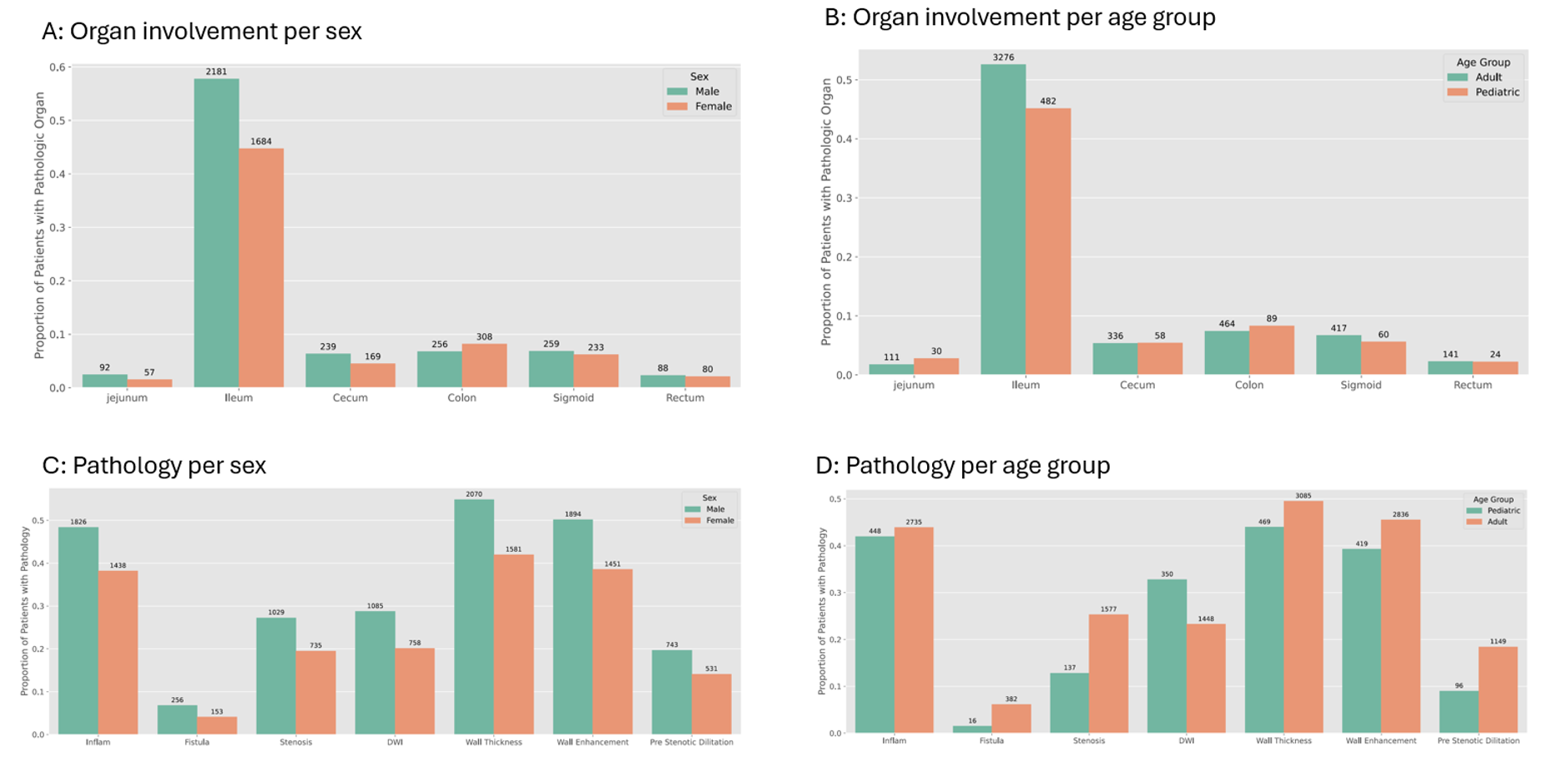}
    \caption{Distribution of organ involvement and pathology prevalence stratified by sex and age group. (A) Organ involvement by sex shows higher ileal involvement in males (2168, 57.78\%) compared to females (1679, 44.80\%), with smaller differences across other gastrointestinal organs. (B) Organ involvement by age group reveals greater ileal and jejunal pathology in adults, while pediatric cases show higher colon and rectal involvement. (C) Pathology prevalence by sex indicates that males have slightly higher rates of inflammation, wall thickness, and wall enhancement, while differences in stenosis and DWI signal are minimal. (D) Pathology by age group reveals higher rates of inflammation and stenosis in pediatric patients, whereas adults exhibit a higher prevalence of pre-stenotic dilatation and wall enhancement.}
    \label{fig:large_scale_analysis}
\end{figure}

Figure~\ref{fig:large_scale_analysis} summarizes the distribution of pathologic organ involvement and key findings stratified by sex and age group. Across the cohort, the ileum was the most affected organ, with higher involvement observed in male (2168, 57.7\%) compared to female (1679, 44.80\%) patients (Figure~\ref{fig:large_scale_analysis}A), and in adults (3264, 52.62\%) compared to pediatric patients (476, 45.08\%) (Figure~\ref{fig:large_scale_analysis}B). Although differences in other organs were less pronounced, the colon and sigmoid showed relatively higher involvement in pediatric patients.
In terms of specific pathologies (Figure~\ref{fig:large_scale_analysis}C and Figure~\ref{fig:large_scale_analysis}D), bowel wall thickening and wall enhancement were among the most prevalent findings overall, particularly in adult patients. Inflammation was more common in males (1816 vs. 1432, 48.40\% vs. 38.21\%) and in adult patients (2723 vs. 444, 43.90\% vs. 42.05\%). Stenosis and pre-stenotic dilatation were more frequent in adults. Notably, DWI signal abnormalities were more common among pediatric patients.

Supplementary Figure~\ref{fig:large_scale_analysis_correlations} presents pairwise correlations between pathological findings and organ involvement across the full dataset. These results reveal clinically meaningful associations, including the frequent co-occurrence of ileal wall thickening with stenosis and pre-stenotic dilatation, and highlight distinct inflammatory profiles between upper (e.g., jejunum, ileum) and lower (e.g., colon, rectum) gastrointestinal segments.

\section{Discussion}
\label{sec:discussion}
We presented HSMP-BERT, a prompt-based framework for extracting structured clinical data from radiology reports, with an emphasis on deployment in low-resource language settings. Our results demonstrate that HSMP-BERT performs well in both zero-shot and fine-tuned configurations. The zero-shot variant generated clinically coherent predictions using only pretrained knowledge, while fine-tuning on a small, manually annotated subset led to further gains—especially for less prevalent findings. These results underscore the value of combining unsupervised pretraining with light supervision for domain-specific NLP tasks.

A key differentiator of HSMP-BERT is its section-aware pretraining via the Section Matching Prediction (SMP) objective. Unlike conventional masked language modeling, SMP explicitly models the logical coherence between the “Findings” and “Impression” sections of radiology reports. This structure-sensitive pretraining equips the model to reason across document segments, improving its utility in downstream clinical extraction tasks. Prompt-based inference further enhances label flexibility and reduces dependence on output-layer retraining.

HSMP-BERT consistently outperformed standard fine-tuning baselines across 24 organ--finding combinations. Performance was particularly strong for the ileum, likely reflecting both the tailored sensitivity of MRE to this region and its relatively balanced label distribution in the dataset. This highlights how anatomical and modality-specific priors embedded in the data can influence model learning.

The hierarchical prompting scheme introduced in our inference pipeline contributed to substantial efficiency gains. By structuring queries from general (e.g., "Is the scan normal?") to specific (e.g., "Is there wall thickening in the ileum?"), we reduced token usage and model calls by over 5-fold. This improvement in computational efficiency is critical for scaling clinical NLP systems to real-world workloads.

Importantly, we demonstrated that HSMP-BERT supports not only per-report classification but also cohort-level analysis. Applied to nearly 10,000 radiology reports, the model identified clinically meaningful patterns across age, sex, and anatomical regions. It revealed co-occurrence structures (e.g., between ileal wall thickening, stenosis, and pre-stenotic dilatation) that align with expected Crohn's disease progression. These findings support the model's potential utility for disease surveillance and cohort characterization.

Compared to prior approaches, HSMP-BERT offers key advantages. Traditional BERT-based models~\cite{Devlin2019,Lee2020,Huang2019} typically treat input as flat sequences and require full fine-tuning on annotated datasets. This limits generalization and impairs performance in zero- or few-shot settings. General-purpose LLMs such as GPT-3~\cite{Kalyan2024} and BioGPT~\cite{Al-Kateb2025} enable prompt-based reasoning but lack domain-aligned pretraining and often fail to capture the structural regularities of radiology text.

By contrast, HSMP-BERT bridges this gap through task-aligned pretraining and structure-guided inference. Its architecture is lightweight, interpretable, and adaptable—attributes that make it especially suited for integration into digital health pipelines where compute and data are constrained.

\subsection{Limitations and Future Work}
\label{sec:limitations}

Despite promising results, several limitations merit discussion. First, fine-tuning remains reliant on annotated data, which can be expensive and time-consuming to obtain in medical domains. Although our approach mitigates this need through SMP pretraining and prompt engineering, broader generalization—especially for rare pathologies—would benefit from additional labeled examples.

Second, HSMP-BERT assumes a semi-structured reporting format (e.g., discrete “Findings” and “Impression” sections). Its performance may degrade on reports that deviate from this convention. Extending the framework to accommodate free-text narratives or loosely structured reports is a promising future direction.

Third, while the model demonstrated strong performance on Hebrew radiology text, further validation is required in other low-resource languages and clinical domains. Releasing a generalized SMP training framework and expanding the ontology of supported labels could enhance cross-setting adaptability.

Finally, although HSMP-BERT is efficient, the hierarchical prompting tree must be manually designed. Automating tree construction or learning task hierarchies from data may improve both scalability and performance.

\section{Conclusion}
\label{sec:conclusion}

In summary, HSMP-BERT offers a robust solution for extracting structured data from unstructured radiology reports. Its design—combining section-aware pretraining, task-specific prompting, and hierarchical inference—enables high performance in low-resource settings with minimal supervision. HSMP-BERT achieves strong results in both zero-shot and fine-tuned configurations, demonstrating versatility across rare and common pathologies. 

Beyond individual report classification, HSMP-BERT supports large-scale applications, including cohort analysis and population-level phenotyping. Its inference efficiency and label generalizability make it well-suited for integration into clinical workflows, retrospective studies, and AI curation pipelines. By bridging the gap between structured data demands and unstructured clinical text, HSMP-BERT advances the state of the art in medical NLP for under-resourced healthcare ecosystems.

\section*{Acknowledgments}
The study was sponsored by the Leona M. and Harry B. Helmsley Charitable Trust. The funders had no role in the study design, data collection and analysis, decision to publish, or preparation of the manuscript.

\section*{Data/Code Availability}
The dataset used in this study contains sensitive patient information and cannot be shared publicly due to privacy and confidentiality regulations. The code will be released upon acceptance.

\bibliographystyle{ieeetr}
\bibliography{hsmp_bert}

\newpage

\section*{Supporting information}
\setcounter{table}{0}
\renewcommand{\thetable}{S\arabic{table}}
\renewcommand{\theHtable}{S\arabic{table}}

\setcounter{figure}{0}
\renewcommand{\thefigure}{S\arabic{figure}}
\renewcommand{\theHfigure}{S\arabic{figure}}

\begin{table}[h!]
\centering
\begin{threeparttable}
\caption{Label prevalence for each of the selected labels (with at least 15 positive examples among the IBD cases). Note: Some labels are near-perfectly balanced (e.g., ``Ileum Wall Thickness''), but most suffer from extreme label imbalance with $<5\%$ positive examples in the annotated dataset..}
\label{tab:supp_label_prevalence}
\begin{tabular}{@{}p{0.62\linewidth}p{0.28\linewidth}@{}}
\toprule
\textbf{Label} & \textbf{Prevalence} \\
\midrule
Ileum Wall Thickness            & 45.2\% (215/476) \\
Ileum Inflammation              & 41.8\% (199/476) \\
Ileum Wall Enhancement          & 34.7\% (165/476) \\
Ileum Stenosis                  & 24.2\% (115/476) \\
Ileum Pre Stenotic Dilatation   & 17.9\% (85/476) \\
Colon Inflammation              & 15.5\% (74/476) \\
Ileum DWI Signal                & 14.7\% (70/476) \\
Ileum Comb sign                 & 14.5\% (69/476) \\
Sigmoid Inflammation            & 13.7\% (65/476) \\
Colon Wall Thickness            & 13.4\% (64/476) \\
Ileum Mesenteric Edema          & 12.8\% (61/476) \\
Sigmoid Wall Thickness          & 11.6\% (55/476) \\
Colon Wall Enhancement          & 8.8\%  (42/476) \\
Rectum Inflammation             & 8.0\%  (38/476) \\
Sigmoid Wall Enhancement        & 7.4\%  (35/476) \\
Ileum Fistula                   & 7.1\%  (34/476) \\
Cecum Inflammation              & 7.1\%  (34/476) \\
Rectum Wall Thickness           & 6.7\%  (32/476) \\
Cecum Wall Thickness            & 6.7\%  (32/476) \\
Cecum Wall Enhancement          & 4.8\%  (23/476) \\
Rectum Wall Enhancement         & 4.4\%  (21/476) \\
Ileum Reduced Motility          & 3.4\%  (16/476) \\
Sigmoid Comb sign               & 3.4\%  (16/476) \\
Sigmoid Mesenteric Edema        & 3.2\%  (15/476) \\
\bottomrule
\end{tabular}
\end{threeparttable}
\end{table}

\begin{table}[t]
\centering
\caption{Performance metrics of the SMP-BERT model across the 24 organ–finding combinations. NPV = negative predictive value, PPV = positive predictive value.}
\label{tab:s2a_metrics}
\begin{adjustbox}{max width=\textwidth}
\begin{tabular}{@{}lcccccc@{}}
\toprule
\textbf{Label} & \textbf{Accuracy} & \textbf{F1 Score} & \textbf{Cohen's $\kappa$} & \textbf{Balanced Accuracy} & \textbf{NPV} & \textbf{PPV} \\
\midrule
Cecum Inflammation              & 0.96 & 0.84 & 0.67 & 0.85 & 0.98 & 0.67 \\
Cecum Wall Enhancement          & 0.97 & 0.83 & 0.65 & 0.78 & 0.98 & 0.80 \\
Cecum Wall Thickness            & 0.96 & 0.84 & 0.68 & 0.84 & 0.98 & 0.70 \\
Colon Inflammation              & 0.94 & 0.88 & 0.77 & 0.90 & 0.97 & 0.78 \\
Colon Wall Enhancement          & 0.94 & 0.79 & 0.58 & 0.77 & 0.96 & 0.67 \\
Colon Wall Thickness            & 0.93 & 0.85 & 0.69 & 0.83 & 0.95 & 0.79 \\
Ileum Comb sign                 & 0.91 & 0.80 & 0.60 & 0.77 & 0.93 & 0.76 \\
Ileum DWI Signal                & 0.97 & 0.93 & 0.87 & 0.91 & 0.97 & 0.95 \\
Ileum Fistula                   & 0.98 & 0.93 & 0.87 & 0.99 & 1.00 & 0.79 \\
Ileum Inflammation              & 0.87 & 0.87 & 0.73 & 0.88 & 0.94 & 0.79 \\
Ileum Mesenteric Edema          & 0.84 & 0.65 & 0.30 & 0.65 & 0.91 & 0.38 \\
Ileum Pre Stenotic Dilatation   & 0.94 & 0.91 & 0.81 & 0.91 & 0.97 & 0.83 \\
Ileum Reduced Motility          & 0.98 & 0.78 & 0.56 & 0.70 & 0.98 & 1.00 \\
Ileum Stenosis                  & 0.95 & 0.93 & 0.87 & 0.95 & 0.98 & 0.86 \\
Ileum Wall Enhancement          & 0.94 & 0.93 & 0.86 & 0.94 & 0.98 & 0.87 \\
Ileum Wall Thickness            & 0.92 & 0.92 & 0.85 & 0.93 & 0.97 & 0.87 \\
Rectum Inflammation             & 0.94 & 0.83 & 0.66 & 0.89 & 0.99 & 0.59 \\
Rectum Wall Enhancement         & 0.96 & 0.80 & 0.61 & 0.84 & 0.99 & 0.66 \\
Rectum Wall Thickness           & 0.97 & 0.88 & 0.77 & 0.94 & 0.99 & 0.69 \\
Sigmoid Comb sign               & 0.97 & 0.63 & 0.27 & 0.60 & 0.97 & 0.52 \\
Sigmoid Inflammation            & 0.87 & 0.75 & 0.51 & 0.79 & 0.95 & 0.52 \\
Sigmoid Mesenteric Edema        & 0.97 & 0.74 & 0.49 & 0.70 & 0.98 & 0.67 \\
Sigmoid Wall Enhancement        & 0.94 & 0.77 & 0.54 & 0.76 & 0.97 & 0.60 \\
Sigmoid Wall Thickness          & 0.89 & 0.74 & 0.48 & 0.75 & 0.94 & 0.53 \\
\bottomrule
\end{tabular}
\end{adjustbox}
\end{table}

\begin{table}[t]
\centering
\caption{Performance metrics of the SMP\-/BERT model when treated as a zero-shot model (no SMP fine-tuning), across the same 24 organ–finding combinations. NPV = negative predictive value, PPV = positive predictive value.}
\label{tab:s2b_metrics}
 \resizebox{\textwidth}{!}{%
\begin{tabular}{@{}lcccccc@{}}
\toprule
\textbf{Label} & \textbf{Accuracy} & \textbf{F1 Score} & \textbf{Cohen's $\kappa$} & \textbf{Balanced Accuracy} & \textbf{NPV} & \textbf{PPV} \\
\midrule
Cecum Inflammation              & 0.52 & 0.39 & -0.01 & 0.49 & 0.93 & 0.07 \\
Cecum Wall Enhancement          & 0.61 & 0.42 & 0.01  & 0.53 & 0.96 & 0.05 \\
Cecum Wall Thickness            & 0.66 & 0.46 & 0.03  & 0.54 & 0.94 & 0.08 \\
Colon Inflammation              & 0.68 & 0.57 & 0.18  & 0.63 & 0.90 & 0.26 \\
Colon Wall Enhancement          & 0.68 & 0.52 & 0.12  & 0.63 & 0.94 & 0.15 \\
Colon Wall Thickness            & 0.76 & 0.61 & 0.23  & 0.65 & 0.91 & 0.29 \\
Ileum Comb sign                 & 0.60 & 0.49 & 0.06  & 0.55 & 0.88 & 0.18 \\
Ileum DWI Signal                & 0.49 & 0.44 & 0.06  & 0.56 & 0.89 & 0.17 \\
Ileum Fistula                   & 0.72 & 0.52 & 0.12  & 0.64 & 0.96 & 0.13 \\
Ileum Inflammation              & 0.53 & 0.51 & 0.02  & 0.51 & 0.59 & 0.43 \\
Ileum Mesenteric Edema          & 0.63 & 0.48 & 0.03  & 0.53 & 0.88 & 0.15 \\
Ileum Pre Stenotic Dilatation   & 0.67 & 0.58 & 0.19  & 0.62 & 0.87 & 0.29 \\
Ileum Reduced Motility          & 0.77 & 0.44 & -0.06 & 0.40 & 0.96 & 0.00 \\
Ileum Stenosis                  & 0.59 & 0.51 & 0.03  & 0.52 & 0.76 & 0.27 \\
Ileum Wall Enhancement          & 0.61 & 0.57 & 0.15  & 0.58 & 0.71 & 0.44 \\
Ileum Wall Thickness            & 0.62 & 0.60 & 0.21  & 0.61 & 0.64 & 0.59 \\
Rectum Inflammation             & 0.59 & 0.38 & -0.11 & 0.36 & 0.89 & 0.02 \\
Rectum Wall Enhancement         & 0.65 & 0.43 & -0.01 & 0.48 & 0.95 & 0.04 \\
Rectum Wall Thickness           & 0.77 & 0.46 & -0.05 & 0.46 & 0.93 & 0.03 \\
Sigmoid Comb sign               & 0.61 & 0.42 & 0.03  & 0.61 & 0.98 & 0.05 \\
Sigmoid Inflammation            & 0.65 & 0.51 & 0.07  & 0.56 & 0.93 & 0.17 \\
Sigmoid Mesenteric Edema        & 0.71 & 0.45 & 0.02  & 0.56 & 0.97 & 0.04 \\
Sigmoid Wall Enhancement        & 0.65 & 0.47 & 0.04  & 0.56 & 0.94 & 0.09 \\
Sigmoid Wall Thickness          & 0.79 & 0.57 & 0.15  & 0.59 & 0.91 & 0.22 \\
\bottomrule
\end{tabular}
}
\end{table}

\begin{table}[t]
\centering
\caption{Performance metrics of a standard fine-tune (SFT) BERT model—identical to the HSMP-BERT backbone—across the same 24 organ–finding combinations. NPV = negative predictive value, PPV = positive predictive value.}
\label{tab:s2c_metrics}
\resizebox{\textwidth}{!}{%
\begin{tabular}{@{}lcccccc@{}}
\toprule
\textbf{Label} & \textbf{Accuracy} & \textbf{F1 Score} & \textbf{Cohen's $\kappa$} & \textbf{Balanced Accuracy} & \textbf{NPV} & \textbf{PPV} \\
\midrule
Cecum Inflammation              & 0.93 & 0.00 & 0.00 & 0.50 & 0.93 & 0.00 \\
Cecum Wall Enhancement          & 0.96 & 0.00 & 0.00 & 0.50 & 0.96 & 0.00 \\
Cecum Wall Thickness            & 0.94 & 0.00 & 0.00 & 0.50 & 0.94 & 0.00 \\
Colon Inflammation              & 0.89 & 0.53 & 0.47 & 0.69 & 0.90 & 0.77 \\
Colon Wall Enhancement          & 0.92 & 0.25 & 0.23 & 0.57 & 0.92 & 1.00 \\
Colon Wall Thickness            & 0.89 & 0.45 & 0.40 & 0.65 & 0.90 & 0.78 \\
Ileum Comb sign                 & 0.87 & 0.41 & 0.35 & 0.64 & 0.89 & 0.64 \\
Ileum DWI Signal                & 0.92 & 0.67 & 0.63 & 0.76 & 0.92 & 0.92 \\
Ileum Fistula                   & 0.93 & 0.15 & 0.14 & 0.54 & 0.94 & 0.50 \\
Ileum Inflammation              & 0.88 & 0.87 & 0.76 & 0.89 & 0.99 & 0.78 \\
Ileum Mesenteric Edema          & 0.87 & 0.09 & 0.06 & 0.52 & 0.88 & 0.33 \\
Ileum Pre Stenotic Dilatation   & 0.90 & 0.70 & 0.64 & 0.80 & 0.92 & 0.76 \\
Ileum Reduced Motility          & 0.97 & 0.00 & 0.00 & 0.50 & 0.97 & 0.00 \\
Ileum Stenosis                  & 0.91 & 0.82 & 0.76 & 0.88 & 0.94 & 0.82 \\
Ileum Wall Enhancement          & 0.91 & 0.87 & 0.80 & 0.91 & 0.97 & 0.81 \\
Ileum Wall Thickness            & 0.91 & 0.90 & 0.81 & 0.91 & 0.99 & 0.83 \\
Rectum Inflammation             & 0.91 & 0.00 & -0.02 & 0.49 & 0.92 & 0.00 \\
Rectum Wall Enhancement         & 0.96 & 0.00 & 0.00 & 0.50 & 0.96 & 0.00 \\
Rectum Wall Thickness           & 0.93 & 0.00 & -0.01 & 0.50 & 0.94 & 0.00 \\
Sigmoid Comb sign               & 0.97 & 0.00 & 0.00 & 0.50 & 0.97 & 0.00 \\
Sigmoid Inflammation            & 0.90 & 0.38 & 0.35 & 0.62 & 0.90 & 1.00 \\
Sigmoid Mesenteric Edema        & 0.97 & 0.00 & 0.00 & 0.50 & 0.97 & 0.00 \\
Sigmoid Wall Enhancement        & 0.93 & 0.00 & 0.00 & 0.50 & 0.93 & 0.00 \\
Sigmoid Wall Thickness          & 0.89 & 0.10 & 0.08 & 0.52 & 0.89 & 0.50 \\
\bottomrule
\end{tabular}
}
\end{table}

\begin{figure}[t!]
    \centering
    \includegraphics[width=0.9\textwidth]{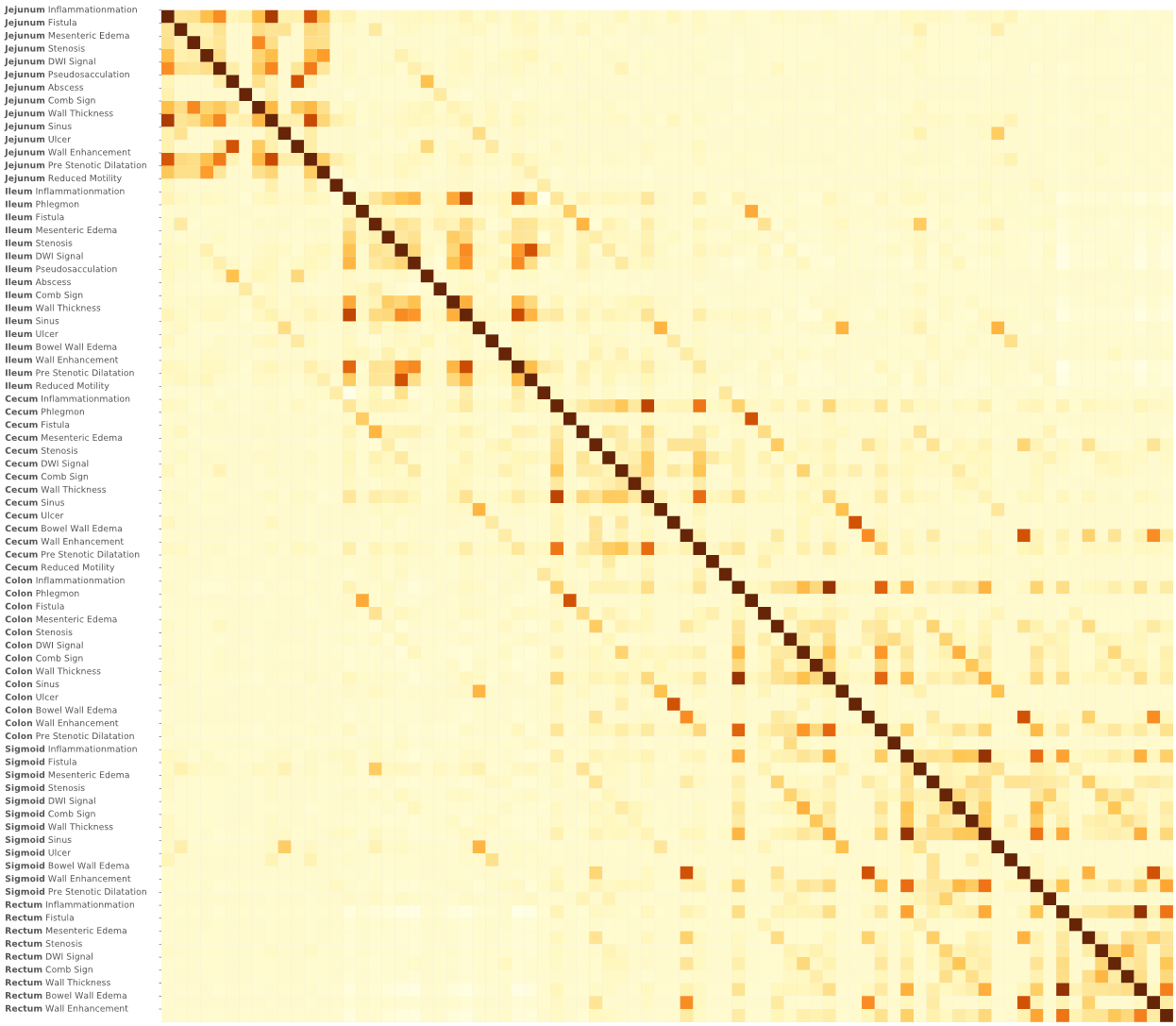}
    \caption{Each cell represents the pairwise Pearson correlation coefficient between the presence of two findings across patients. Rows and columns are ordered by anatomical location, from proximal (jejunum) to distal (rectum), facilitating interpretation of spatial anatomical patterns.}
    \label{fig:large_scale_analysis_correlations}
\end{figure}

\end{document}